\documentclass[runningheads]{llncs}
\usepackage[T1]{fontenc}
\usepackage{graphicx}
\usepackage{csquotes}
\usepackage{tikz}
\usepackage{comment}
\usepackage{amsmath,amssymb} 
\usepackage{color}
\usepackage{wrapfig}
\usepackage{booktabs}
\usepackage{multirow}
\usepackage{multicol}
\usepackage{pifont}
\usepackage[ruled,vlined]{algorithm2e}
\newcommand{\cmark}{\ding{51}}%
\newcommand{\xmark}{\ding{55}}%
\usepackage[misc]{ifsym}
\usepackage[accsupp]{axessibility}  

\begin{document}
\title{Make A Long Image Short: Adaptive Token Length for Vision Transformers}
%
%
\author{Qiqi Zhou \inst{1}\textsuperscript{,*} \and
Yichen Zhu \inst{2}\textsuperscript{,\Letter}}
%
\authorrunning{Q. Zhou, Y. Zhu.}
%
\institute{Shanghai University of Electric Power, Shanghai, China \and
Midea Group, Shanghai, China\
\\
\email{\{zhouqq31, zhuyc25\}@midea.com}
\footnote[0]{\textsuperscript{*}~Work done during internships at Midea Group.
\\ \textsuperscript{\Letter}~Corresponding authors.}
}
%
\maketitle              
\begin{abstract}
The vision transformer is a model that breaks down each image into a sequence of tokens with a fixed length and processes them similarly to words in natural language processing. Although increasing the number of tokens typically results in better performance, it also leads to a considerable increase in computational cost. Motivated by the saying "A picture is worth a thousand words," we propose an innovative approach to accelerate the ViT model by shortening long images. Specifically, we introduce a method for adaptively assigning token length for each image at test time to accelerate inference speed. First, we train a Resizable-ViT (ReViT) model capable of processing input with diverse token lengths. Next, we extract token-length labels from ReViT that indicate the minimum number of tokens required to achieve accurate predictions. We then use these labels to train a lightweight Token-Length Assigner (TLA) that allocates the optimal token length for each image during inference. The TLA enables ReViT to process images with the minimum sufficient number of tokens, reducing token numbers in the ViT model and improving inference speed. Our approach is general and compatible with modern vision transformer architectures, significantly reducing computational costs. We verified the effectiveness of our methods on multiple representative ViT models on image classification and action recognition.
\end{abstract}
\section{Introduction}
The transformer has achieved remarkable success in computer vision since the introduction of ViT~\cite{dosovitskiy2020image}. It has demonstrated impressive performance compared to convolutional neural networks (CNNs) on various visual domains, including image classification~\cite{deit,twins}, object detection~\cite{carion2020end,zhu2020deformable}, semantic segmentation~\cite{liu2021swin}, and action recognition~\cite{fan2021mvit,timesformer}, using both supervised and self-supervised~\cite{he2021masked,bao2021beit} training configurations. Despite the development of ViT models, their deployment remains a challenge due to the high computational cost associated with them.

\begin{wrapfigure}{R}{0.5\textwidth}
    \centering
    \includegraphics[width =0.5\textwidth]{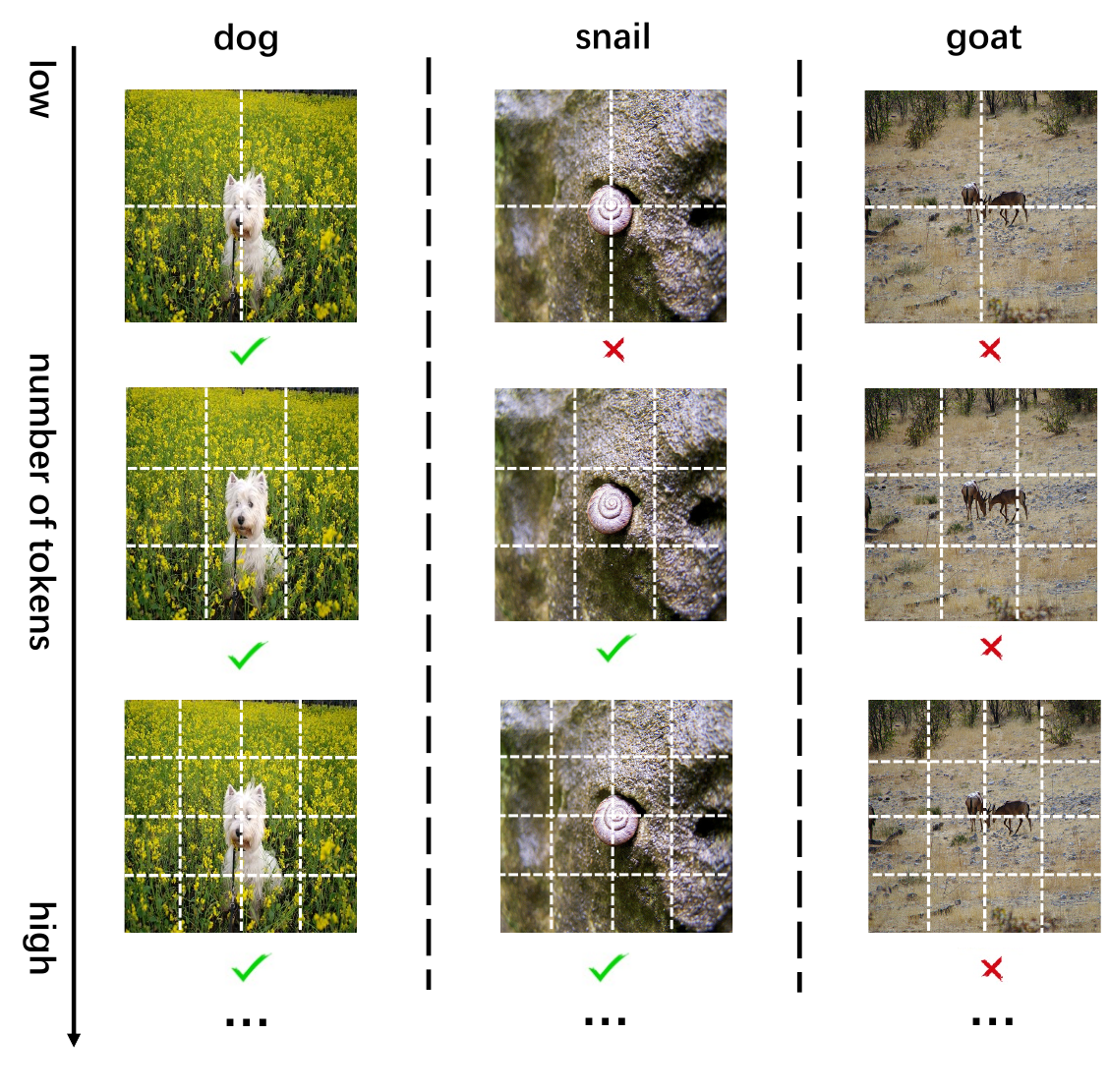}\\
    \vspace{-3 mm}
      \caption{The motivation for our approach. While some images (right) may need many tokens to predict their category, some images are easy to recognize. Thus, only a small number of tokens is sufficient to classify them correctly.}\label{fig:motivation}

\end{wrapfigure}

Accelerating ViT is a crucial yet understudied area. While many techniques like pruning, distillation, and neural architecture search have been applied to accelerate CNNs, these cannot be directly applied to ViT due to significant differences between the models~\cite{naseer2021intriguing,mahmood2021robustness,raghu2021vision}. As the attention module in the transformer computes the fully-connected relations among all input patches~\cite{vaswani2017attention}, the computational cost becomes quadratic with respect to the length of the input sequence~\cite{choromanski2020rethinking,beltagy2020longformer}. Consequently, the transformer can be computationally expensive, particularly for longer input sequences. In the ViT model, images are divided into a fixed number of tokens; following conventional practice~\cite{dosovitskiy2020image}, an image is represented by $16 \times 16$ tokens. We aim to reduce the computational complexity of ViT by reducing the number of tokens used to split the images. Our motivation is depicted in Figure~\ref{fig:motivation}, which shows three examples predicted by individually trained DeiT-S models~\cite{deit} with different token lengths. The checkmark denotes correct prediction, and the cross denotes the wrong prediction. We observe that some "easy-to-classify" images only require a few tokens to determine their category accurately, while some images require more tokens to make the right prediction. These observations motivate us to reduce the computational complexity of the existing ViT model by accurately classifying the input using the minimum possible number of tokens.

In an ideal scenario, we would know the minimum number of tokens required to accurately predict an image, and we could train a model to assign the optimal token length to the ViT model. However, training multiple ViT models, each with a fixed token length, would be computationally infeasible. To address this, we propose a modification to the transformer architecture, changing it from "static" to "dynamic," enabling the ViT model to adaptively process images with varying token lengths. This dynamic transformer, called Resizable-ViT (ReViT), identifies the minimum token length required to achieve correct predictions for each image. We then train a lightweight Token-Length Assigner (TLA) to predict the appropriate token length for a given image, with the label obtained from the ReViT. Consequently, the ReViT can process images with lower computational costs based on the assigned token length.

The primary challenge of our approach is training the ReViT to enable the ViT model to process images of any size provided by the TLA. To tackle this challenge, we introduce a token length-aware layer normalization that switches the normalization statistics for each type of token length, and a self-distillation module that enhances the model's performance when using short token lengths in ReViT. Additionally, the ViT model needs to see the images with the corresponding token lengths beforehand to handle various token lengths effectively. However, as the number of predefined token-length choices increases, the training cost linearly increases. To overcome this, we introduce a parallel computing strategy for efficient training that makes the ReViT training almost as inexpensive as a vanilla ViT model's training.

We showcase the efficacy of our approach on several prominent ViT models, such as DeiT~\cite{deit} and LV-ViT~\cite{lvvit} for image classification, and TimesFormer~\cite{timesformer} for video recognition. Our experiments demonstrate that our method can significantly reduce computational costs while maintaining performance levels. For instance, we achieve a 50\% acceleration in DeiT-S~\cite{deit} model with an accuracy reduction of only 0.1\%. On action recognition, the computational cost of TimesFormer~\cite{timesformer} can be reduced up to 33\% on Kinetic 400 with only a 0.5\% loss in recognition accuracy.

\section{Related Works}
\noindent
\textbf{Vision transformer.} ViT have recently gained much attention in computer vision due to their strong capability to model long-range relations. Many attempts have been made to integrate long-range modeling into CNNs, such as non-local networks~\cite{wang2018non,yin2020disentangled}, relation networks~\cite{hu2018relation}, among others. Vision Transformer (ViT)\cite{dosovitskiy2020image} introduced a set of pure Transformer backbones for image classification, and its follow-ups have soon modified the vision transformer to dominate many downstream tasks for computer vision, such as object detection\cite{carion2020end,zhu2020deformable}, semantic segmentation~\cite{liu2021swin}, action recognition~\cite{timesformer,fan2021mvit}, 2D/3D human pose estimation~\cite{yang2020transpose,poseformer}, 3D object detection~\cite{pointformer}, and even self-supervision~\cite{he2021masked}. ViT has shown great potential to be an alternative backbone for convolutional neural networks.
\\
\\
\noindent
\textbf{Dynamic vision transformer.} The over-parameterized model is known to have many attractive merits and can achieve better performance than smaller models. However, in real-world scenarios, computational efficiency is critical as executed computation is translated into power consumption or carbon emission. To address this issue, many works have attempted to reduce the computational cost of Convolutional Neural Networks (CNNs) through methods such as neural architecture search~\cite{liu2018darts,cai2019once,zoph2016neural}, knowledge distillation~\cite{zhu2021student,zhao2022decoupled,hinton2015distilling,scalekd,undistill}, and pruning~\cite{han2015deep,frankle2018lottery}. 

Recent work has shift its attention to reduce the number of tokens used for inference, as the number of tokens can be a computational bottleneck to the vision transformer. There are two major approaches: unstructured token sparsification and structured token division. The majority of works, including PatchSlim~\cite{tang2021patch}, TokenSparse~\cite{rao2021dynamicvit}, GlobalEncoder~\cite{song2021dynamic}, IA-RED~\cite{pan2021ia}, and Tokenlearner~\cite{ryoo2021tokenlearner}, focus on the former. TokenLearner~\cite{ryoo2021tokenlearner} uses an MLP to reduce the number of tokens. TokenPooling~\cite{marin2021tokenpool} merges tokens via a k-mean based algorithm. TokenMerge~\cite{bolya2023tokenmerge} calculates the token similarity and merges tokens via bipartite soft matching.
.

They aim to remove uninformative tokens, such as those that learn features from the background of the image, thereby boosting inference speed by reserving only informative tokens. These approaches typically need to progressively reduce the number of tokens based on the inputs and can be performed either jointly with ViT training or afterward. However, pruning tokens sparsely can bring unstable training issues, especially when the model is huge~\cite{li2022scaling}. 

The latter, which is known as unstructured token sparsification, is the most relevant work to our research. Wang et al.\cite{wang2021not} proposed Dynamic Vision Transformer (DVT) to dynamically determine the number of patches required to divide an image. They employed a cascade of ViT models, with each ViT responsible for a specific token length. The cascade ViT model makes a sequential decision and stops inference for an input image if it has sufficient confidence in the prediction at the current token length. In contrast to DVT\cite{wang2021not}, our method is more practical and accessible, as it only requires a \textit{single} ViT model. Additionally, we focus on how to \textit{accurately} determine the minimum number of token lengths required in the transformer to provide correct predictions for each image.

\section{Methodology}
The vision transformers treat an image as a sentence by dividing the 2D image into 1D tokens and modeling the long-range dependencies between them using the multi-head self-attention mechanism. However, the self-attention is considered the computational bottleneck in the transformer model, as its computational cost increases quadratically with the number of incoming tokens. As mentioned earlier, our approach is motivated by the observation that many \enquote{easy-to-recognize} images do not require $16\times16$ tokens~\cite{dosovitskiy2020image} to be correctly classified. Therefore, computational costs can be reduced by processing fewer tokens on \enquote{easy} images while using more tokens on \enquote{hard} images. It is worth noting that the key to a successful input-dependent token-adaptive ViT model is to determine precisely the minimum number of tokens required to accurately classify the image.

\begin{figure*}[t]
    \begin{minipage}{1.0\linewidth}
    \centering
    \includegraphics[width =0.8\textwidth]{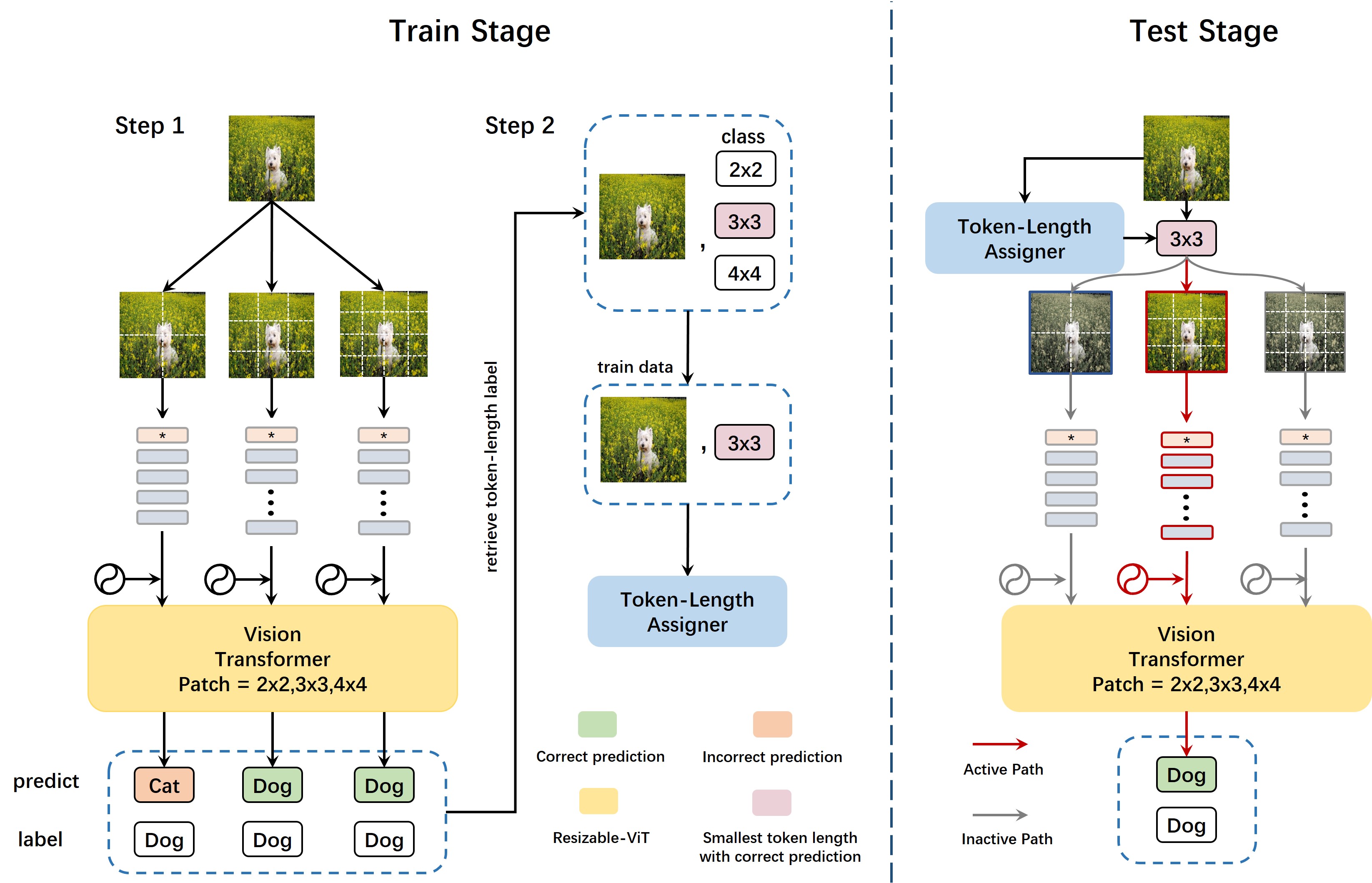}\\
    \end{minipage}
    \caption{\textbf{Left}: There are two steps in the training procedure. First, we train the Resizable-ViT that can split an image into any predefined token length. Secondly, we train a Token-Length Assigner based on the token-length label that is retrieved from ReViT. It is the smallest number of tokens that can correctly predicate the class of the image. \textbf{Right}: In inference, the TLA first assigns a token-length for the image, then ReViT uses this setting to make predication.}\label{fig:overview}
\end{figure*}

To achieve our goal, we propose a two-stage model training approach. In the first stage, we train a ViT model that can handle images with any predefined token lengths. Usually, a single ViT model can only handle one token length. We describe the model design and training strategy of this ViT model in detail in Section~\ref{sec:revit}. In the second stage, we train a model to determine the appropriate token length for each image. We first obtain the token-length label, which represents the minimum number of tokens required for accurate classification, from the previously trained ViT model. Then, we train a Token-Length Assigner (TLA) using the training data, where the input is an image and the label is the corresponding token length. This decoupled procedure allows the TLA to make a better decision regarding the number of tokens required for each image. During inference, the TLA guides the ViT model on the optimal number of tokens required for accurate classification based on the input. The complete training and testing process is illustrated in Figure~\ref{fig:overview}.

In the following, we first introduce the Token-Label Assigner, then present the training method on the Resizable-ViT model and improved techniques.

\subsection{Token-Length Assigner}
The purpose of the Token-Length Assigner (TLA) is to make accurate predictions based on the feedback from ReViT. TLA training is performed after ReViT. We first define a list of token lengths $L = [l_{1}, l_{2}, \dots, l_{n}]$ in descending order. For simplicity, we use a single number to represent the token length, such as $L = [14 \times 14, 10 \times 10, 7 \times 7]$. The model with a token length of $7 \times 7$ has the lowest computational cost among the three token lengths.

\begin{figure*}[!ht]
    \begin{minipage}{1.0\linewidth}
    \centering
    \includegraphics[width =0.65\textwidth]{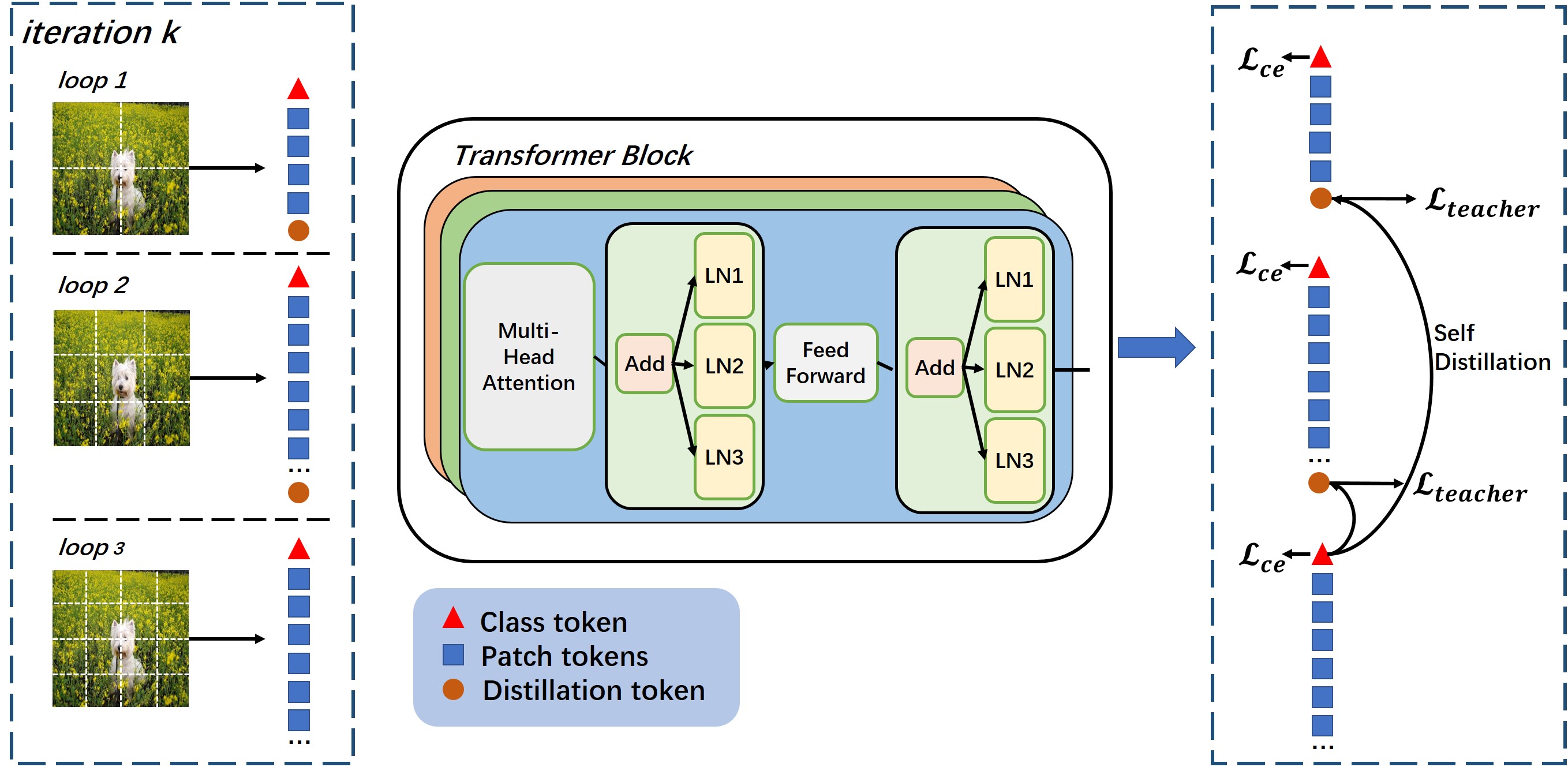}\\
    \end{minipage}
    \vspace{-3 mm}
      \caption{Example of self-distillation and token-length aware layer normalization in ReViT. Each token length corresponds to a LayerNorm (LN in this figure) and pass-through this LayerNorm during both training and inference. The self-distillation is only conducted in training, where smaller token lengths have an extra distillation token to learn from the teacher's knowledge.}\label{fig:betterstudetn}
\end{figure*}

In order to train a token-length adapter (TLA), it is necessary to obtain a token-length label from the ReViT model at convergence. For an image, the token-length label is defined as the minimum token length required by the ViT model to accurately classify that image. The inference speed of the ReViT model, denoted by $M$, can be ranked as $Speed(M_{l_{1}}) < Speed(M_{l_{2}}) < \dots < Speed(M_{l_{k}})$, where $k = len(L)$ represents the total number of options for token length. For each input $x$, we can obtain the prediction $y_{l_{i}} = M_{l_{i}}(X)$ for all $i \in n$. The label of the input $x$ is determined by the smallest token size $l_{j}$ for which any smaller token length would result in an incorrect prediction, i.e., $y_{l_{j-1}} \neq y^{gt}$, where $gt$ is the ground truth label. Therefore, a set of input-output pairs $(x, l_{j})$ can be obtained and used to train the TLA. Since token-label assignment is straightforward, the TLA is a lightweight module, with minimal computational overhead introduced. Moreover, since unnecessary tokens are reduced in the ViT model, the additional computational overhead is relatively small.

\subsection{Resizable-ViT} \label{sec:revit}
In this section, we present the Resizable-ViT (ReViT), a dynamic ViT model capable of accurately classifying images with various token lengths. We introduce two techniques that enhance the performance of ReViT and subsequently present the training strategy. Additionally, we offer an efficient training implementation that accelerates the training process of ReViT.
\\
\\
\noindent
\textbf{Token-Aware Layer Normalization.}
The Layer Normalization (LN/LayerNorm) layer is a widely used normalization technique that accelerates training and improves the generalization of the Transformer architecture. In both natural language processing and computer vision, it is common to adopt an LN layer after addition in the transformer block. However, as the feature maps of the self-attention matrices and feed-forward networks constantly change, the number of token sizes changes as well. Consequently, inaccurate normalization statistics across different token lengths are shared in the same layer, which impairs test accuracy. Additionally, we found empirically that LN cannot be shared in ReViT.

To address this issue, we propose a Token-Length-Aware LayerNorm (TAL-LN), which uses an independent LayerNorm for each choice of token length in the predefined token length list. In other words, we use $Add$ $\&$ $\{LN_{1}$, ..., $LN_{k}\}$ as a building block, where $k$ represents the number of predefined token lengths. Each LayerNorm layer calculates layer-wise statistics specifically and learns the parameters of the corresponding feature map. Furthermore, the number of extra parameters in TAL-LN is negligible since the number of parameters in normalization layers typically takes less than one percent of the total model size~\cite{yu2018slimmable}. A brief summary is illustrated in Figure~\ref{fig:betterstudetn}.
\\
\\
\noindent
\begin{algorithm}[t]
\SetAlgoLined
\textbf{Require:} Define Token-Length Assigner $T$, token-length list $\mathbf{R}$, for example, $\{16, 24, 32\}$. The iterations $N_M$ for training $M$. The $CE(\cdot)$ denotes cross-entropy loss, and $DisT(\cdot)$ denotes distillation loss.\\
\For{\upshape $t = 1, \dots, N_{M}$} {
    \upshape
    Get data $x$ and class label $y_{c}$ of current mini-batch.\\
    Clear gradients for all parameters, \textit{optimizer.zero\_grad()}\\
    \For{\upshape $i = 1, \dots, len(\mathbf{R}) - 1$}
    {
    Convert ReviT to selected token-length $M_{i}$,  \\
    Execute current scaling configuration. $\hat{y}_{i} = M_{i}(x)$.\\
    \If{$\mathbf{R}[i]$ == 16}{
      set teacher label. $\hat{y}_{i}^{teacher} =\hat{y}_{i}$ \\
      Compute loss \textit{$loss_{i} = CE(\hat{y}_{i}, y)$} \\
    }\Else{
      Compute loss \textit{$loss_{i} = DisT(\hat{y}_{i}^{teacher}, \hat{y}_{i}, y)$} \\
    }
    Compute gradients, \textit{$loss_{i}.backward()$} \\
    }
    Update weights, \textit{$optimizer.step()$.} \\
    }
Obtain token-length label for all train data $(x, y_{t})$.\\
Train $T$ with $(x, y_{t})$.\\
\caption{Training Resizable-ViT $M$.}
\label{alg:alg1}
\end{algorithm}
\textbf{Self-Distillation}
It is aware that the performance of ViT is strongly correlated to the number of patches, and experiments have shown that reducing the token size significantly hampers the accuracy of small token ViT. Directly optimizing via supervision from the ground truth poses a challenge for the small token length sub-model. Motivated by self-attention, a variant of knowledge distillation techniques, where the teacher can be insufficiently trained, or even the student model itself~\cite{yu2018slimmable,yu2019universally,yu2020bignas}, we propose a token length-aware self-distillation (TLSD). In the next section, we will show that the model with the largest token length $M_{1}$ is always trained first. For $M_{l_{1}}$, the training objective is to minimize the cross-entropy loss $\mathcal{L}{CE}$. When it comes to the model with other token lengths $M{l_{i}}, i \leq k, i \neq 1$, we use a distillation objective to train the target model:
\begin{equation}
    \mathcal{L}_{teacher} = (1 - \lambda)\mathcal{L}_{CE}(\phi(Z_{s}), y) + \lambda \tau^2KL(\phi(Z_{s}/\tau), \phi(Z_{t}/\tau))
\end{equation}
where $Z_{s}$ and $Z_{t}$ is the logits of the student model $M_{l_{i}}$ and teacher model $M_{l_{1}}$, respectively. $\tau$ is the temperature for the distillation, $\lambda$ is the coefficient balancing the KL loss (Kullack-Leibler divergence) and the CE loss (cross-entropy) on ground truth label $y$, and $\phi$ is the softmax function. Similar to DeiT, we add a distillation token for student models. Figure~\ref{fig:betterstudetn} gives an overview. Notably, this distillation scheme is computational-free: we can directly use the predicted label of the model with the largest token length as the training label for other sub-model, while for the largest token length model, we use ground truth.

\subsection{Training Strategy}

\begin{wrapfigure}{h}{0.5\textwidth}
    \centering
    \includegraphics[width =0.5\textwidth]{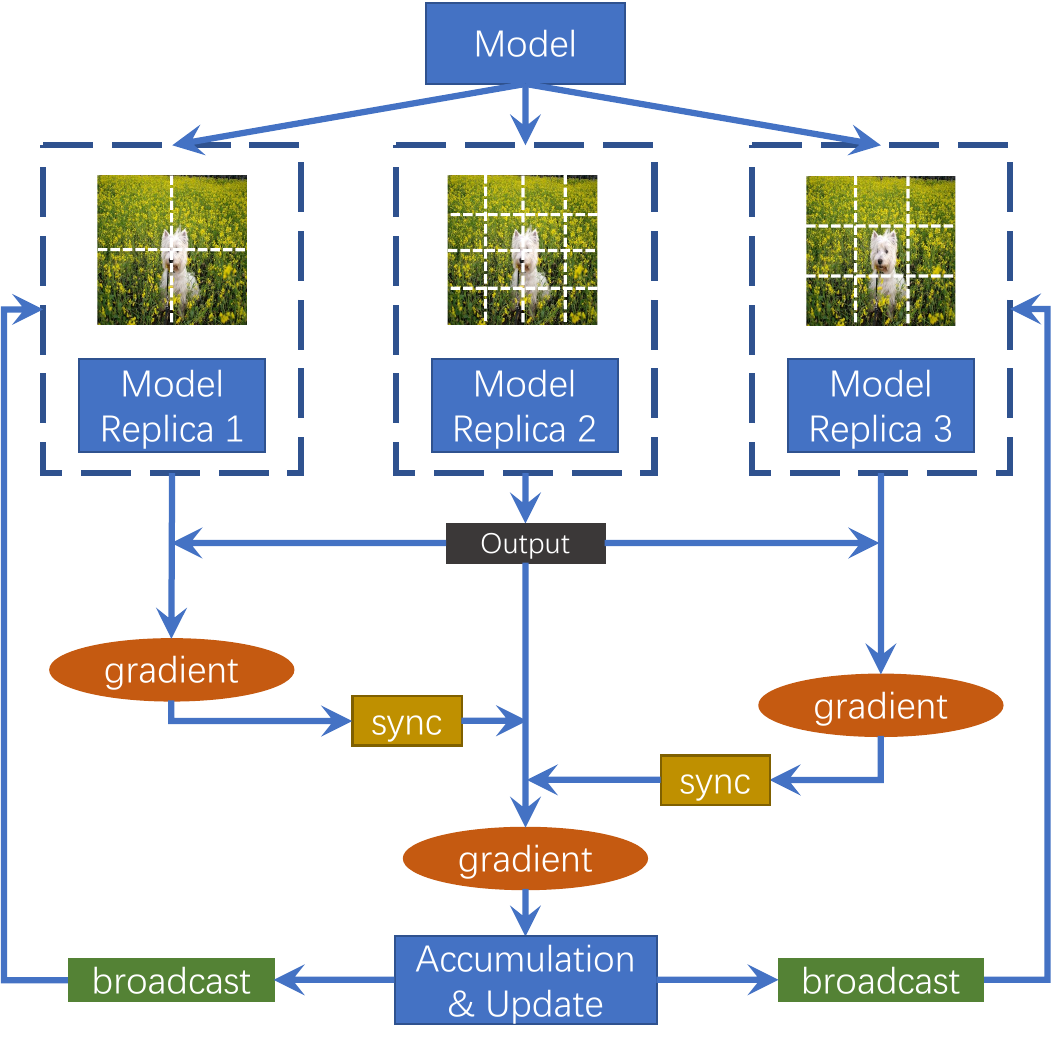}\\
      \caption{Efficient training implement for Resizable Transformer through parallel computing. All gradient from the replicate nodes are synchronize on the node where that have the largest token length to save the cost of communication.}\label{fig:parallel}
\end{wrapfigure}

To enable the ViT model to adaptively process various token lengths in the predefined choice list, it is necessary to expose it to images with different token lengths. Inspired by batch gradient accumulation, a technique used to overcome the problem of small batch size by accumulating gradient and batch statistics in a single iteration, we propose a mixing token length training. As shown in Algorithm~\ref{alg:alg1}, a batch of images is processed with different token lengths to compute the loss through feed-forward, and individual gradients are obtained. After looping through all token length choices, the gradients of all parameters calculated by feeding different token lengths are accumulated to update the parameters.

\noindent
\textbf{Efficient Training Implementation.} 
An issue with the aforementioned training strategy is that the training time increases linearly with the number of predefined token length choices. To address this issue, we propose an efficient implementation strategy that trades memory cost for training time. As shown in Figure~\ref{fig:parallel}, we replicate the model, with each model corresponding to a specific token length. At the end of each iteration, the gradients of the different replicas are synchronized and accumulated. Notably, we always send the gradient of replicas in which the token length is small to the one with a larger token length, as they are the training bottleneck. Thus, the communication cost in the gradient synchronization step is negligible. Then, the model parameters are updated through back-propagation. After the parameter updating is complete, the main process distributes the learned parameters to the rest of the replicas. These steps are repeated until the end of training, after which all replicas except the model in the main process can be removed. As such, the training time of the Resizable Transformer reduces from $O(k)$ to $O(1)$, where $k$ is the number of predefined token lengths. Though the number of $k$ is small, i.e., $k=3$, in practice, the computational cost of training $k$ ViT is high. Through our designed parallel computing, the training cost for ReViT is almost the same as that of naive ViT, where the cost of communication between replicas is negligible compared to the model training cost. In exchange for fast training, extra computational power is required for parallel computing.

\begin{figure*}[t]
    \begin{minipage}{1.0\linewidth}
    \centering
    \includegraphics[width =0.8\textwidth]{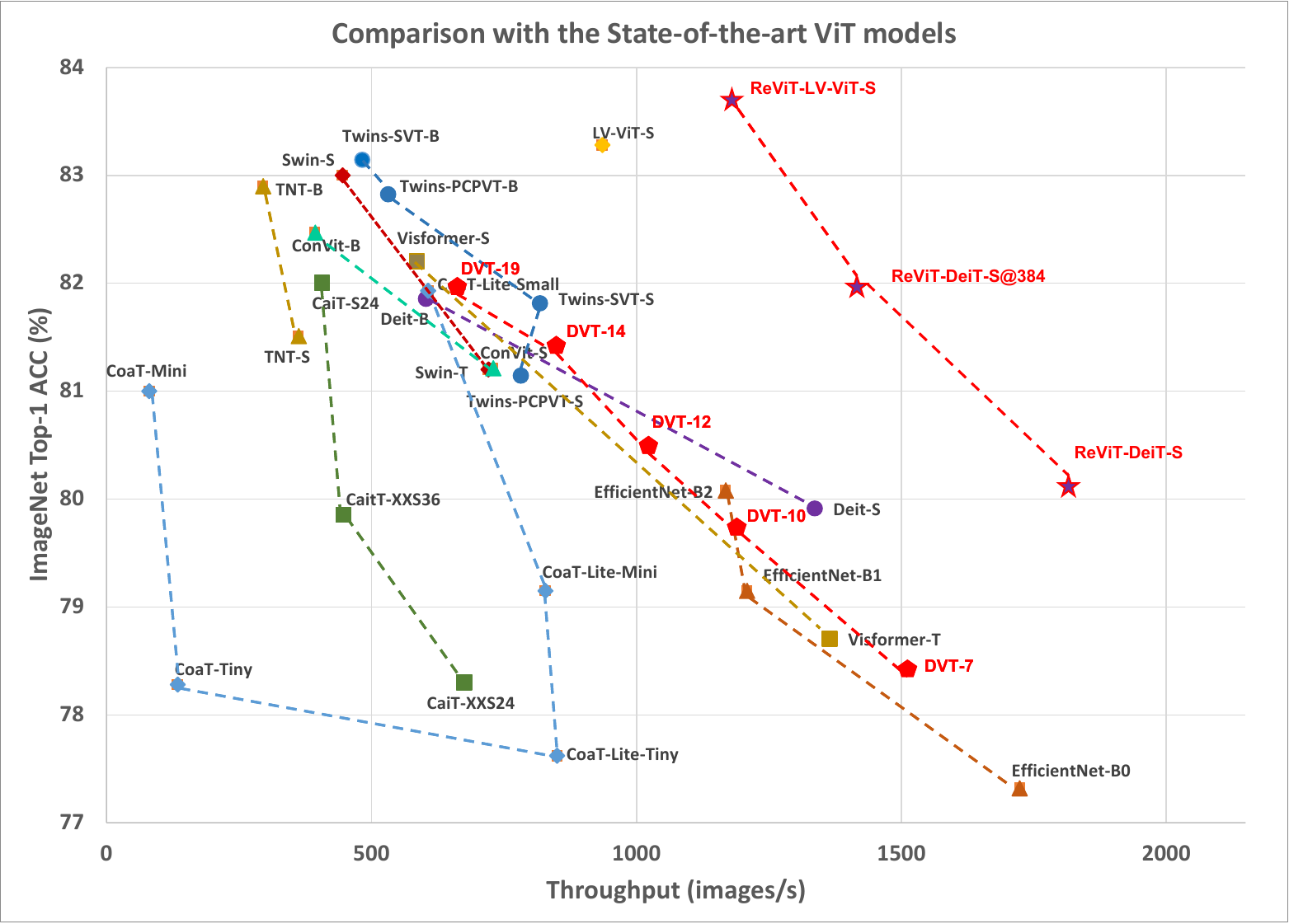}\\
    \end{minipage}
    \vspace{-3 mm}
      \caption{Comparison of different models with various accuracy-throughput trade-off. The throughput is measured on an NVIDIA RTX 3090 GPU with batch size fixed to 32. The input image size is $224 \times 224$ unless indicate otherwise. The ReViT (red stare in the figure) achieves better trade-off than other methods, including DVT~\cite{wang2021not}.}\label{fig:overall_perm}
\end{figure*}

\section{Experiments}
\noindent
\textbf{Implementation details.} For image classification, we trained all models on the ImageNet~\cite{deng2009imagenet} training set consisting of around 1.2 million images and reported their accuracy on the 50k test images. The predefined token lengths were set to $14 \times 14$, $10 \times 10$, and $7 \times 7$ by default, with the token length of $4 \times 4$ excluded due to a significant accuracy drop. We conducted experiments on DeiT-S~\cite{deit} and LV-ViT-S~\cite{lvvit} using an image resolution of 224 $\times$ 224, unless otherwise specified. We followed the training settings and optimization methods described in the original papers of DeiT~\cite{deit} and LV-ViT~\cite{lvvit}. For LV-ViT, we obtained token labels for smaller token lengths using their proposed method. We also trained the ReViT on resized images with higher resolutions, such as 384 on DeiT-S. To avoid optimization difficulties caused by large kernel and stride convolutional layers required for patch embedding, we replaced them with consecutive convolutions followed by the method in Xiao et al.~\cite{xiao2021early}. After training the ReViT, we obtained token-length labels for all training data and trained the Token-Length Assigner (TLA), which was a small version of EfficientNet-B0 compared to the ViT model. We also included feature map transfer and attention transfer as part of self-distillation, which we found empirically useful. 
We use Something-Something V2~\cite{goyal2017something} to conduct experiments on action recognition. The Something-Something V2
is another large-scale video dataset, having around 169k videos for training and 20k videos for validation. We follow the training setting of MotionFormer~\cite{patrick2021motionformer}. Specifically, two versions of MotionFormer are tested. The default version operates on $16 \times 224 \times 224$ video clips, and a high spatial resolution variant operates on $32 \times 448 \times 448$ video clips. 

\begin{table*}[t]
  \caption{\textbf{Video recognition on Something-Something V2.} Our ReViT outperforms state-of-the-art CNN-based and ViT-based methods. IN-21 and K400 are abbreviations for ImageNet-21K and Kinetic-400 datasets.}
  \label{tbl:video_exp}
  \centering
  \renewcommand*\arraystretch{1.1}
  \resizebox{0.95\textwidth}{!}{\begin{tabular}{l|c|c|c|c|c|c}
    \hline
    \hline
     \textbf{Method} & \textbf{Backbone} & \textbf{FLOPs (G)} & \textbf{Top-1 ($\%$)} & \textbf{Top-5 ($\%$)} &  \textbf{Frames} & \textbf{Extra Data}\\
    \hline
    TEINet~\cite{liu2020teinet} & ResNet50 & 99$\times$10 $\times$3 &  66.5 & - & \multirow{3}{*}{8+16} & \multirow{3}{*}{ImageNet-1K} \\
    TANet~\cite{tanet}  & ResNet50 & 99 $\times$ 2 $\times$ 3 & 66.0 & 90.1 &  &\\
    TDN~\cite{wang2021tdn} & ResNet101 & 198$\times$1$\times$ 3 & 69.6 & 92.2 & & \\
    \hline
    SlowFast~\cite{feichtenhofer2019slowfast} & ResNet101 &106$\times$1$\times$3 & 63.1 & 87.6 & 8+32 & \multirow{2}{*}{Kinetics-400} \\
    MViTv1~\cite{fan2021mvit} & MViTv1-B & 455$\times$1$\times$3 & 67.7 & 90.9 & 64 & \\
    \hline
    TimeSformer~\cite{timesformer} & ViT-B & 196$\times$1$\times$3 & 59.5 & - & 8 & \multirow{2}{*}{ImageNet21K} \\
    TimeSformer~\cite{timesformer} & ViT-L & 5549$\times$1$\times$3 & 62.4 & - & 64 & \\
    \hline
    ViViT~\cite{arnab2021vivit} & ViT-L  & 995$\times$4$\times$3 & 65.9 & 89.9 & 32 & \multirow{4}{*}{IN-21K + K400} \\
    Video Swin~\cite{liu2022videoswin} & Swin-B & 321$\times$1$\times$3 & 69.6 & 92.7 & 32 &  \\
    Motionformer~\cite{patrick2021motionformer} & ViT-B & 370$\times$1$\times$3 & 66.5 & 90.1 & 16 &  \\
    Motionformer~\cite{patrick2021motionformer} & ViT-L & 1185$\times$1$\times$3 & 68.1 & 91.2 & 32 & \\
    \hline
    \textbf{ReViT}\_{\text{motionformer}} & ViT-B & 183$\times$1$\times$3 & 66.6 & 89.9 & 16 & \multirow{2}{*}{IN-21K + K400} \\
    \textbf{ReViT}\_{\text{motionformer}} & ViT-L & 570$\times$1$\times$3 & 67.6 & 90.8 & 32 & \\ 
    \hline
    \hline
  \end{tabular}}
\end{table*}
\subsection{Experimental Results}
\textbf{Main Results on ImageNet Classification.} We present the main results of our ReViT based on DeiT-S and LV-ViT-S in Figure~\ref{fig:overall_perm}. Our approach is compared with several models, including DeiT~\cite{deit}, CaiT~\cite{cait}, LV-ViT~\cite{lvvit}, CoaT~\cite{coat}, Swin~\cite{liu2021swin}, Twins~\cite{twins}, Visformer~\cite{visformer}, ConViT~\cite{wu2021cvt}, TNT~\cite{tnt}, and EfficientNet~\cite{tan2019efficientnet}. The results show that our method achieves a favorable accuracy-throughput trade-off. Specifically, ReViT reduces the computational cost of the baseline counterpart by decreasing the token number used for inference. By increasing the input resolution, we manage to outperform the baseline counterpart, given a similar computational cost. We also highlight the experimental results of DVT~\cite{wang2021not} in red. Our method achieves significantly better performance in terms of both accuracy and throughput. We hypothesize that despite the low FLOPs of DVT, the practical speed of DVT is high due to its multiple cascade ViT structure.
\\
\\
\noindent
\textbf{Main Results on Video Recognition.} One of the core motivations behind ReViT is to address the issue of high computational costs in extremely long token lengths during inference for image classification tasks. To further explore this idea, we investigate the applicability of our method to video recognition tasks, where the token length in transformers is typically much longer than that in image classifiers.

\begin{figure*}[t]
    \centering
    \includegraphics[width =0.9\textwidth]{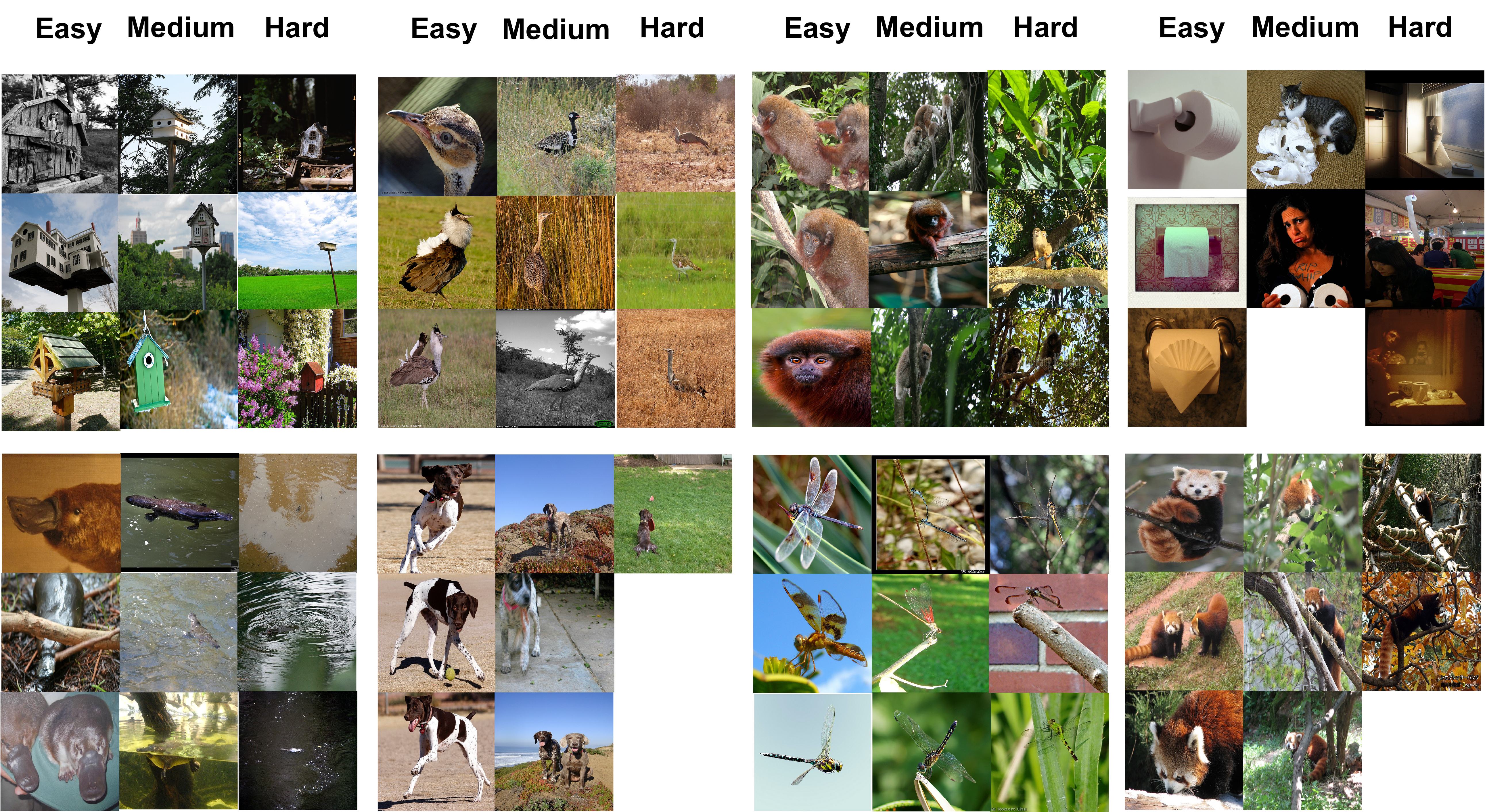}
    \vspace{-3 mm}
      \caption{Visualization of \enquote{hard}, \enquote{medium}, and \enquote{easy} samples that predicted by Token-Length Assigner and which the ReViT-DeiT-S got correction prediction. Most of the \enquote{easy} images have clear sight on the object, while size of objects is mostly small for \enquote{hard} samples.}\label{fig:visualization}
\end{figure*}

To this end, we train the ReViT-MotionFormer models with ViT-B and ViT-L, two different backbones, and compare them with the baseline models, respectively. The results are presented in Table~\ref{tbl:video_exp}. Our method demonstrates a significant speedup over the MotionFormer baseline, with a computational cost reduction of approximately 51\% and a 0.1\% accuracy increase. By training on larger image resolutions, we correspondingly reduce the model size by 48\% with a 0.5\% accuracy drop, which is slightly worse than the smaller resolution counterpart. Nonetheless, our experiments demonstrate that ReViT is effective for action recognition tasks.
\\
\\
\noindent
\textbf{Visualization of samples with different token-length.} 
\label{sec:visual} We selected eight classes from the ImageNet validation set and chose three samples from each category, classified as easy, medium, and hard, corresponding to tokens with dimensions of $14 \times 14$, $10 \times 10$, and $7 \times 7$, respectively. The image samples were selected based on the token length assigned by the Token-Length Assigner. The resulting images are displayed in Figure~\ref{fig:visualization}. Notably, some classes do not have all images filled because less than three samples in the validation set belong to those categories. For example, only one image in the dog class requires the largest token length for classification. We observe that the number of tokens required to predict the category is highly correlated with the object's size. For larger objects, only a few tokens are sufficient to predict their category.

\begin{table}[t]
\centering
\caption{The ablation study of self-distillation in ReviT. The SD* denotes the self-distillation. We also evaluate the performance with different choices of $\tau$. The self-distillation improves the performance notably, the small token length model outperforms the baseline when $\tau = 0.9$.}
\begin{tabular}{l|c|c|c|c|c}
\toprule
\multirow{2}{*}{Method} & \multirow{2}{*}{SD*} & \multirow{2}{*}{$\tau$}  & \multicolumn{3}{c}{Top-1 Acc (\%)} \\
   &       &      &     14 $\times$ 14 & 10 $\times$ 10 & 7 $\times$ 7 \\
\midrule
Deit-S  & \xmark    &   -    &  79.85 & 74.68  & 72.41  \\
\midrule
\multirow{3}{*}{ReViT} &   \xmark  &  -  & 80.12 & 74.24 & 70.15\\
& \cmark & 0.5 &  79.92 & 76.16  & 71.33 \\
& \cmark & 0.9 & 79.83  &  76.86 & 74.21 \\
\bottomrule
\end{tabular}
\label{tbl:self_distill}
\end{table}

\begin{table}[t]
\centering
\caption{The ablation study of shared patch embedding and shared position encoding in ReViT. The Pos denotes the positional encoding module. We notice that sharing these two modules decrease the model accuracy.}
\begin{tabular}{l|c|c|c|c|c}
\toprule
\multirow{2}{*}{Method} & \multicolumn{2}{c}{Shared}   & \multicolumn{3}{c}{Top-1 Acc (\%)} \\
   &    Patch   &   Pos   &     14 $\times$ 14 & 10 $\times$ 10 & 7 $\times$ 7 \\
\midrule
\multirow{3}{*}{ReViT} &   \xmark  &  \cmark  & 65.14 &  61.30 &  58.35\\
& \cmark & \xmark & 75.24 & 71.32  & 69.73 \\
& \cmark & \cmark & 79.83  &  76.85 & 74.21 \\
\bottomrule
\end{tabular}

\label{tbl:sharep}
\end{table}
\subsection{Ablation Study}
\noindent
\textbf{Shared patch embedding and position encoding.}
We conducted an experiment to evaluate the impact of using shared patch embedding and position encoding. As the token number changes during training, we applied some techniques to enable sharing of both operations. To handle position encoding, we followed the approach of ViT~\cite{dosovitskiy2020image} and zero-padded the position encoding module whenever the token size changed. This technique was initially used to adjust the positional encoding in the pretrain-finetune paradigm. For shared patch embedding, we used a weight-sharing kernel~\cite{cai2019once}. A large kernel was constructed to process a large patch size, and when the patch size changed, a smaller kernel with shared weight on the center was adopted to flatten the image patch.

As shown in Table~\ref{tbl:sharep}, both shared patch embedding and shared positional encoding decreased the model's accuracy. In particular, the accuracy dropped by nearly 14\% for the large token length model when using the shared patch strategy. The shared positional encoding module performed better than shared patch embedding but still significantly impacted the performance of ReViT.
\\
\\
\noindent
\textbf{The effect of self-distillation and choice of $\tau$.} \label{sec:selfdistill_exp} We conducted experiments to verify the effectiveness of self-distillation in ReViT and investigated the impact of the hyper-parameter $\tau$. We tested two different values of $\tau$, 0.9 and 0.5, for all sub-networks and demonstrated the results in Table~\ref{tbl:self_distill}. Without self-distillation, the accuracy on small token lengths was comparable to tokens of size $10 \times 10$, but significantly worse on tokens of size $7 \times 7$. When we applied self-distillation with $\tau = 5$, the accuracy of both models increased. To further evaluate the model, we used $\tau = 5$. The higher value of $\tau$ negatively impacted the accuracy of the largest token length, dropping the accuracy by around 0.3\%, but significantly improving the performance of models with token size $7 \times 7$. This highlights the necessity of using self-distillation in our scenario and demonstrates the importance of carefully selecting the hyper-parameter $\tau$ for optimal performance.
\begin{figure}[t]
    \begin{minipage}{1.0\linewidth}
    \centering
    \includegraphics[width =\textwidth]{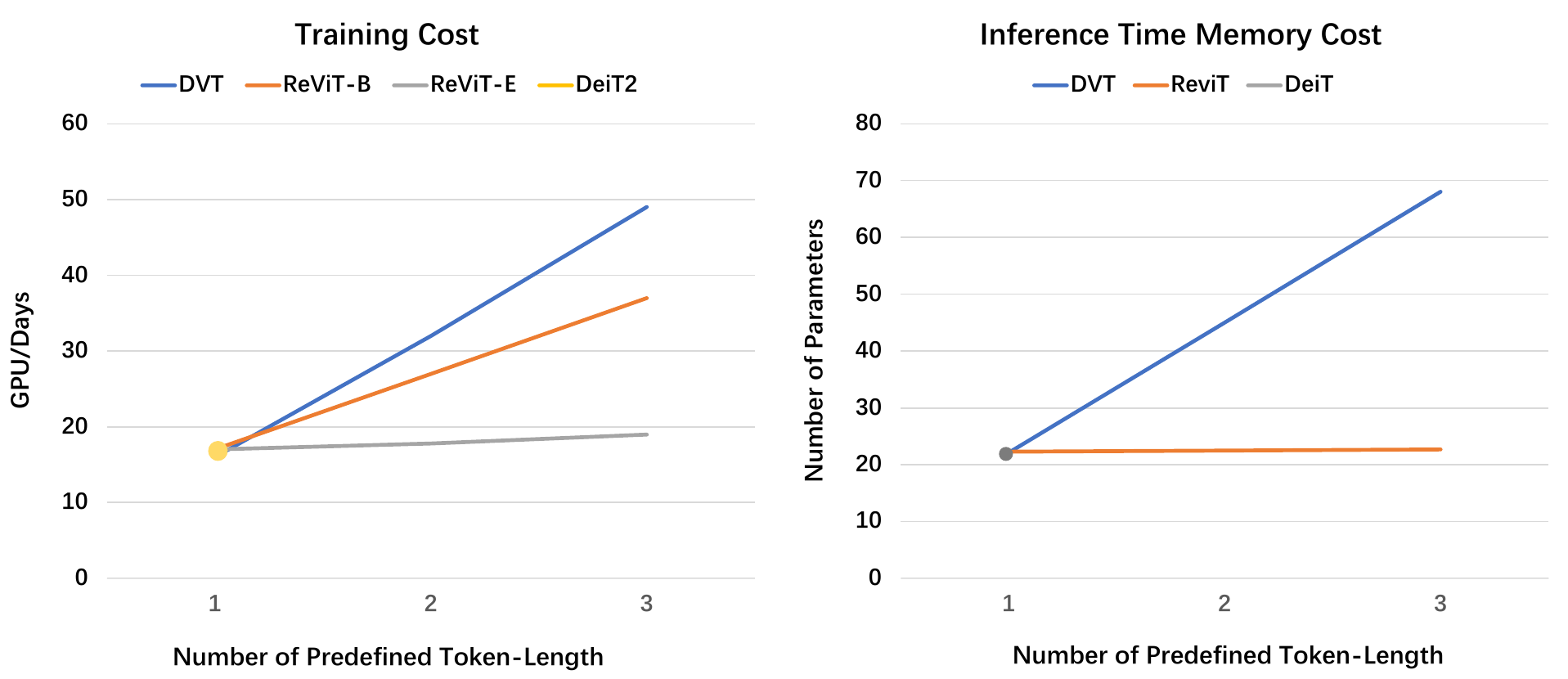}\\
    \end{minipage}
    \vspace{-3 mm}
      \caption{Compare our approach with DeiT-S~\cite{deit} and DVT~\cite{wang2021not} for training cost and memory cost at inference time in terms of the number of predefined token-length. Our proposed ReViT is almost a cheap as training the baseline DeiT-S, while DVT requires linearly increased budget on training and memory.}\label{fig:cos_ana}
\end{figure}
\\
\\
\textbf{Training cost and Memory Consumption.} \label{sec:traincost} We compared ReViT with DeiT-S and DVT~\cite{wang2021not} in terms of training cost and memory consumption, as shown in Figure~\ref{fig:cos_ana}. ReViT-B denotes the baseline approach of ReViT, while ReViT-E is the efficient implementation method. Both ReViT-B and DeiT-S show a linear increase in training cost as the number of choices in $s$ increases. ReViT-B is cheaper because backpropagation of multiple token lengths is merged. However, the training time of ReViT-E slightly increases due to the communication cost between parallel models increasing.

As for memory consumption (number of parameters) during testing, since our method only has a single ViT where most computational heavy components are shared, the memory cost is slightly higher than the baseline. However, compared to DVT, the increase in the number of parameters with respect to the increasing number of token length choices is negligible. This indicates that our approach is more practical than DVT in terms of both training cost and memory cost. Furthermore, our method is easier to apply to existing ViT models than DVT.
\begin{figure*}[t]
    \begin{minipage}{1.0\linewidth}
    \centering
    \includegraphics[width =0.6\textwidth]{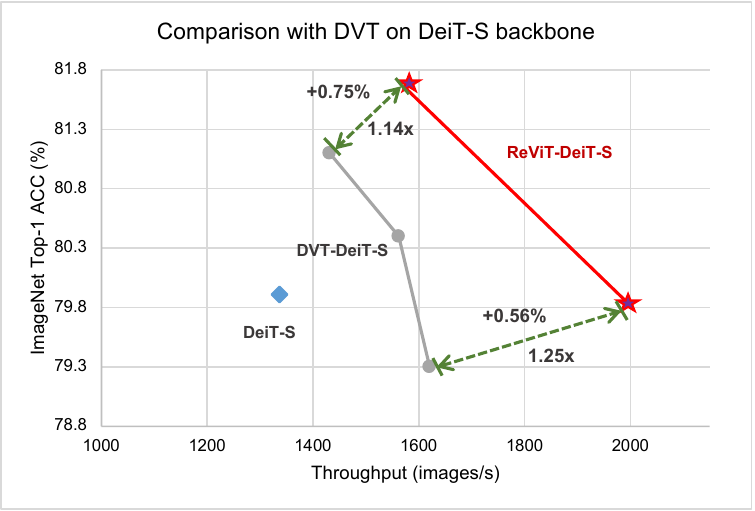}\\
    \end{minipage}
    \vspace{-3 mm}
      \caption{Comparison with DVT~\cite{wang2021not} on DeiT-S backbone. Our method outperforms DVT by a large margin.}\label{fig:dvt_ablation}
\end{figure*}
\\
\\
\noindent
\textbf{Comparison with DVT.} We conducted a further investigation of our proposed method based on DeiT-S and compared it with DVT, which was also developed based on DeiT-S. Figure~\ref{fig:dvt_ablation} shows that our proposed ReViT achieves superior performance compared to DVT. This could be due to our better selection of the number of patches that achieves the best accuracy-speed tradeoff.

\section{Conclusions}
This paper aims to reduce the token length to split the image in the ViT model to eliminate unnecessary computational costs. First, we propose the Resizable Transformer (ReViT), which adaptively processes any predefined token size for a given image. Then, we define a Token-Length Assigner to decide the minimum number of tokens that the transformer can use to classify the individual image correctly. Extensive experiments indicate that ReViT can significantly accelerate the state-of-the-art ViT model. Also, compared to the prior SOTA method, our approach achieves better training speed, inference cost, and model performance. Therefore, we believe our paper benefits practitioners who would like to adopt ViT in deployment.

%
%
\newpage
\bibliographystyle{splncs04}
\bibliography{egbib}

\end{document}